\documentclass{article}

\usepackage{microtype}
\usepackage{doi}
\usepackage{graphicx}
\usepackage{subfigure}
\usepackage{booktabs}
\usepackage{nth}% for professional tables
\usepackage{listings}
\PassOptionsToPackage{hyphens}{url}\usepackage{hyperref}
\usepackage{xurl}

\usepackage[accepted]{icml2025}
\usepackage{amsmath}
\usepackage{amssymb}
\usepackage{mathtools}
\usepackage{amsthm}
\usepackage[capitalize,noabbrev]{cleveref}

\theoremstyle{plain}

\theoremstyle{definition}

\theoremstyle{remark}
\usepackage{xcolor}
\usepackage{tikz}
\usepackage{lipsum}
\usepackage[most]{tcolorbox}

\newtcolorbox{personalstatement}{colback=gray!10,enhanced,boxrule=0pt,after skip=0.5cm,before skip=0.5cm,right skip=0cm,breakable,fonttitle=\bfseries,toprule=0pt,bottomrule=0pt,rightrule=0pt,leftrule=4pt,arc=0mm,colframe=darkgray,colbacktitle=darkgray,boxed title style={
		frame code={ 
			\fill[darkgray](frame.south west)--(frame.north west)--(frame.north east)--([xshift=3mm]frame.east)--(frame.south east)--cycle;
		}
	}
}

\usepackage[textsize=tiny]{todonotes}

\icmltitlerunning{Reproducibility: The New Frontier in AI Governance}

\begin{document}

\twocolumn[
\icmltitle{Reproducibility: The New Frontier in AI Governance}

\icmlsetsymbol{equal}{*}

\begin{icmlauthorlist}
\icmlauthor{Israel Mason-Williams}{yyy}
\icmlauthor{Gabryel Mason-Williams}{comp}
\end{icmlauthorlist}

\icmlaffiliation{yyy}{UKRI Safe and Trusted AI, Imperial College London and King's College London, London, United Kingdom}
\icmlaffiliation{comp}{Independent Researcher, London, United Kingdom}

\icmlcorrespondingauthor{Israel Mason-Williams}{israel.mason-williams@kcl.ac.uk}

\icmlkeywords{Machine Learning, ICML}

\vskip 0.3in
]

\printAffiliationsAndNotice{}  

\begin{abstract}
AI policymakers are responsible for delivering effective governance mechanisms that can provide safe, aligned and trustworthy AI development. However, the information environment offered to policymakers is characterised by an unnecessarily low Signal-To-Noise Ratio, favouring regulatory capture and creating deep uncertainty and divides on which risks should be prioritised from a governance perspective. We posit that the current publication speeds in AI combined with the lack of strong scientific standards, via weak reproducibility protocols, effectively erodes the power of policymakers to enact meaningful policy and governance protocols. Our paper outlines how AI research could adopt stricter reproducibility guidelines to assist governance endeavours and improve consensus on the AI risk landscape. We evaluate the forthcoming reproducibility crisis within AI research through the lens of crises in other scientific domains; providing a commentary on how adopting preregistration, increased statistical power and negative result publication reproducibility protocols can enable effective AI governance. While we maintain that AI governance must be reactive due to AI's significant societal implications we argue that policymakers and governments must consider reproducibility protocols as a core tool in the governance arsenal and demand higher standards for AI research. 
Code to replicate data and figures: \url{https://github.com/IFMW01/reproducibility-the-new-frontier-in-ai-governance}
\end{abstract}

\section{Introduction}

\begin{figure}[htb]
    \centering
    \includegraphics[width=0.72\linewidth]{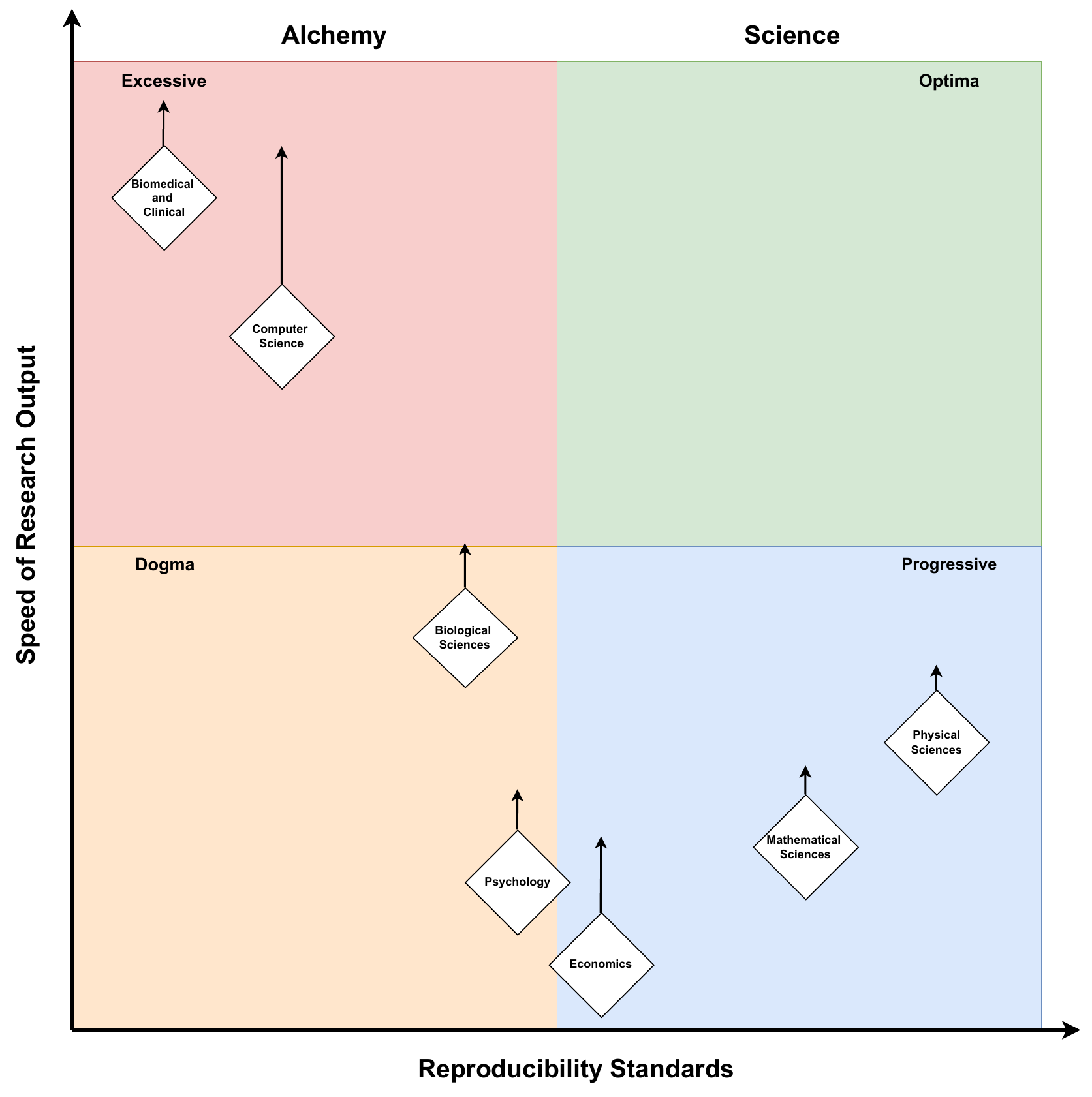}
    \caption{Indicative plot of the speed of publication versus reproducibility standards in scientific domains with average publication trajectories ($\uparrow$) over the last five years. Please see Appendix Section~\ref{sec:Speed_Reproduce} for the methodology used to produce this plot.}
    \label{fig:speed_vs_reproducibility}
\end{figure}

AI is often regarded as a technology that will have an unprecedented impact on technological development with speculated impacts on society including, but not limited to, scientific research advances~\cite{abramson2024accurate,cory2024evolving}, changes to global economics ~\cite{trammell2023economic}, up-ending job markets~\cite{kulveit2025gradual,eloundou2023gpts}, new cyber security threats and opportunities~\cite{dash2022threats}, and revolutionising healthcare~\cite{lee2021application}. With the increasing economic, scientific and societal interest in this multi-purpose technology, many perspectives have emerged on where current trajectories will lead us, with prominent voices arguing both for the imminent arrival of Artificial General Intelligence (AGI)~\cite{grace2024thousands} and against its theoretical plausibility~\cite{van2024reclaiming}. Despite the lack of consensus within the scientific community on the trajectory of AI, due to hype dynamics of AGI, there is increasing pressure for regulators and policymakers to respond to the range of risks and prepare for potential futures offered by AI advancements. Furthermore, given that current AI systems propagate and amplify complex biases~\cite{caliskan2023artificial,10.1145/3582269.3615599}, it is crucial that the Signal-To-Noise Ratio for AI research is increased, to make it easier to assess AI capabilities and better position regulators and policymakers to act on accurate and trustworthy research. 
\begin{personalstatement}
    \textbf{Signal-To-Noise Ratio:} The quantity of research papers that contain genuine/reproducible \textbf{(signal)} insights compared to the number of papers that contain stochastic/irreproducible \textbf{(noise)} findings.
\end{personalstatement}
The current standard of scientific research in AI has led many prominent AI researchers to warn of a reproducibility crisis~\cite{kapoor2023leakage,ball2023ai,gundersen2020reproducibility,gundersen2018state,tran2021an}. The definition of reproducibility is somewhat contested in literature so we explicitly define reproducibility, in line with the B2 definition provided by~\cite{desai2025reproducibility}. It is important to note that the Open Science Collaboration makes no distinction between these terms~\cite{open2012open} and the Association for Computing Machinery used these definitions until later updating them~\footnote{ACM Artifact Review and Badging: \url{https://www.acm.org/publications/policies/artifact-review-and-badging-current}}. Due to the AI communities use of the term reproducibility~\footnote{AI4Europe Reproducibility Initiative: \url{https://www.ai4europe.eu/ethics/articles/ai4europe-reproducibility-initiative}}~\footnote{ML Reproducibility Challenge:~\url{https://reproml.org} } we have opted to center our work under this term to enable the AI community to access this work.
\begin{personalstatement}
    \textbf{Replicability}: An independent group of researchers obtain the same result of a study using independently developed artifacts in-line with the original experimental setup \textbf{(different team, same experimental setup)}.
    \\
    \\
    \textbf{Reproducibility}: An independent group of researchers obtain the same findings of a study using independently developed artifacts under a modified methodological approach and/or dataset \textbf{(different team, different experimental setup)}.
\end{personalstatement}
It is broadly accepted that at the start of the 21st century, many different research domains, such as, Economics, Cancer Biology, and Psychology, experienced such reproducibility crises, the fallout of which has led to ineffective economic policy, opportunity cost, loss of life and ineffective medical treatments. In Figure~\ref{fig:speed_vs_reproducibility} we visualise an \textbf{indicative} plot of scientific domains current publication speed versus reproducibility efforts and their projected growth, as represented in Table~\ref{tab:pub_over_years}. We argue that the trajectory for AI research can be improved with the introduction of strong reproducibility protocols such as \textbf{preregistration, statistical leverage and negative result reporting}. In this paper we contextualise our recommendations and the importance of reproducibility in science by discussing how other domains have dealt with similar reproducibility issues. Through our insights we hope to bolster transparency and trust in AI research such that effective governance strategies for AI can be enacted.

\section{Learning from Past Reproducibility Crises}
Many empirical scientific fields have faced a reproducibility crisis~\cite{christensen2018transparency,errington2021reproducibility,open2015estimating}. In this section we cover some of the landmark reproducibility crises discussing the impacts of poor science and how each field attempted to create more robust reproducibility protocols to mitigate irreproducibility. It is important to understand core case studies for irreproducibility such that we can contextualise the impact that irreproducible AI research can have; as well as the potential harm propagation if better standards are not adopted. Furthermore failure to replicate/reproduce can foster overconfidence, underestimate uncertainty, and hinder scientific progress~\cite{errington2021investigating} and thus, thwart policy and governance efforts.

\subsection{Economics - Reproducibility}

Economic research is an important factor in policy, therefore, it is crucial that rigorous scientific practices are upheld to ensure genuine advances in knowledge are made~\cite{o1992economists} . In 2006 a study on the Journal of Money, Credit, and Banking, which has mandatory data sharing requirements, found that most authors did not adhere to these practices with only 25.56\% (68/266) having archival entries where required~\cite{mccullough2006lessons}. Furthermore, it was found that only 22.58\% (14/62) of empirical studies could be replicated, this study sparked further inquiries into reproducibility in economics. More recent studies in economics  have shown low replicability with successful replication of less than half of a selected group of papers published across top economic journals~\cite{chang2015economics}.
\\
\\
A core case study that typifies the impact of irreproducible findings is found in the ``Growth in The Time of Debt'' paper published in the American Economic Review which explored the systemic relationship between public debt, growth and inflation~\cite{reinhart2010growth}. The paper asserted ``When external debt reaches 60 percent of GDP [Gross Domestic Product], annual growth declines by about two percent; for higher levels, growth rates are roughly cut in half''~\cite{reinhart2010growth}. In the wake of the 2008 financial crash the relation made between external debt and growth had major impacts on governmental perspectives to economic policy. Governments sought austerity policies which seeks to reduce budget deficits, and therefore reliance on external debt, by leveraging spending cuts for public services and/or tax increases. It has been argued that the Eurozone used the evidence to support ``The Treaty on Stability, Coordination and Governance'' which stated that member states should not exceed debt in excess of 60\% of GDP~\cite{compact2012treaty} and that it was used to support austerity policy in the United Kingdom. An Oxfam case study report for the United Kingdom revealed that austerity measures led to an increase in inequality and created an environment for the rich to get richer~\cite{THE} and further studies have linked austerity policies in the United Kingdom to hundreds of thousands of excess deaths~\cite{walsh2022bearing}.

An attempt to replicate the results of the ``Growth in the Time of Debt'' paper failed due to missing data and existing errors in the calculations of the original work~\cite{bell2015stylised}. When correctly analysing the data  the replicators asserted, that there was no trend that the OECD countries conformed with regard to debt and growth. This result meant that there was no evidence to support adopting austerity policy from the original study's findings; demonstrating how over-reliance on specific irreproducible findings can have deep unintended consequences for societies. In an attempt to mitigate the harms of non replicable or reproducible economic research, there has been an increased emphasis on the importance of research design, preregistration, disclosure standards, and open sharing of data and materials~\cite{christensen2018transparency} to improve the transparency and credibility of research outputs in this domain.
\begin{table*}[t]
\centering
\caption{Publication trends across scientific domains over the last five years, as categorised by  Dimensions database as of April 2025.}
\resizebox{\textwidth}{!}{
\begin{tabular}{|c|c|c|c|c|c|c|c|}
\hline
\textbf{Research Domain}                                                       & \textbf{2019} & \textbf{2020} & \textbf{2021} & \textbf{2022} & \textbf{2023} & \textbf{2024} & \textbf{\begin{tabular}[c]{@{}c@{}}Percentage Growth\\ 2019-2024\end{tabular}} \\ \hline
\begin{tabular}[c]{@{}c@{}}Biomedical \& Clinical\end{tabular}             & 1,170,895     & 1,345,291     & 1,417,197     & 1,433,960     & 1,435,700     & 1,478,650     & 26.284\%                                                                     \\ \hline
\begin{tabular}[c]{@{}c@{}}Information \& Computer Science\end{tabular} & 475,933       & 520,807       & 590,753       & 639,524       & 723,629       & 818,642       & 72.008\%                                                                    \\ \hline
Biological                                                                     & 388,231       & 439,843       & 475,523       & 488,074       & 478,308       & 487,150       & 25.479\%                                                                     \\ \hline
Physical                                                                       & 284,936       & 296,182       & 311,913       & 299,285       & 303,834       & 318,808       & 11.888\%                                                                     \\ \hline
Mathematical                                                                   & 187,573       & 197,366       & 203,721       & 208,431       & 207,464       & 212,854       & 13.478\%                                                                     \\ \hline
Psychology                                                                     & 146,967       & 160,994       & 171,286       & 173,419       & 176,259       & 176,589       & 20.156\%                                                                     \\ \hline
Economics                                                                      & 81,421        & 90,407        & 95,953        & 101,201       & 106,581       & 109,335       & 34.284\%                                                                     \\ \hline
\end{tabular}}
\label{tab:pub_over_years}
\end{table*}
\subsection{The Reproducibility Project: Cancer Biology}

In 2021 the Center for Open Science concluded an eight-year-long study to replicate 193 experiments from 53 high-impact preclinical papers in cancer biology published between 2010 and 2012~\cite{errington2021reproducibility}. Preclinical papers provide the foundation for determining which treatments to give further evaluation and testing in clinical trials on humans. While this is an important stage of treatment development, it can represent a large opportunity cost to participants when other known treatment routes are available~\cite{kane2021preclinical}. When considering cancer patients, where time can be limited, this is of particular concern. In general it is reported that 19 of 20 cancer drugs used for clinical studies do not demonstrate enough safety, efficacy or commercial promise to achieve license, which incurs a significant financial and opportunity cost~\cite{kane2021preclinical}. The Open Science reproduction study for cancer biology found that only 2\% of studies had open data, 0\% of the studies had pre-requisite protocols (a detailed plan for conducting a research study) that allowed for replication and that of the experiments that could be successfully reproduced (50/193)  the effect sizes were 85\% smaller on average than the original findings~\cite{errington2021reproducibility}. The alarming results provides a particularly harrowing perspective on the importance for rigorous reproducibility efforts, especially when considering the integration of AI into such domains. Furthermore, the low Signal-To-Noise Ratio means it is challenging to identify ideal cancer drug candidates in the future, pushing back scientific progress. However, it is important to note that cancer will not merely be solved via strong scientific standards alone. Following findings from this report (and other earlier studies) reproducibility recommendations for cancer biology were made that include expert statistician evaluation of experiments, preregistration, preprinting with public comments (to avoid publication bias) and transparent data and code availability~\cite{rodgers2021have}. 

\subsection{Reproducibility Project: Psychology}
In 2012 a study conducted by the Open Science Collaboration (OSC) commenced to examine the reproducibility of psychology research~\cite{open2015estimating}. In the study, the OSC attempted to replicate 100 randomly selected studies from three prestigious psychology journals. In 2015, following a three-year-long project, the results were released. To replicate the studies they used the original materials and high-powered designs. They discovered that only 36\% of the studies ``successfully'' replicated had significance in the same direction as the original studies, but that effect sizes were half that of the effect size reported in the original studies~\cite{open2015estimating}. The study largely pointed towards cultural issues surrounding pressure to publish and argued that incentives for individual scientists prioritise novelty over reproduction. Such studies have prompted the development of the Transparency and Openness Promotion Guidelines~\cite{nosek2016transparency} which introduces a TOP Factor metric that reports a journals alignment with promoting core scholarly norms of transparency and reproducibility~\cite{center2020new}.

\subsection{Reflections for AI}
While AI is regarded as the most important emerging technology~\cite{IEEE} like any other empirical science, it remains vulnerable to reproducibility pitfalls without robust research practices and protocols. Furthermore, it is important to note that given publications in AI have been growing on average at a circa 50\% faster rate between 2019 and 2024 compared to most domains in the last five years, as shown in Table~\ref{tab:pub_over_years}, there is a strong requirement to establish robust scientific practices before the number of publications exceeds a critical threshold where the Signal-To-Noise Ratio is too low. Without intervention increasing the ratio could represent a significant challenge beyond the scope of any individual or group, of researchers, publishing bodies and policymakers. In the absence of consensus on AI reproducibility and research standards, it is possible that industry actors will be able to capitalise on a polluted information environment which creates a threat of regulatory capture through asymmetric information~\cite{baron1984regulation} and could undermine AI governance endeavours and democratic systems.

\section{Mitigation Strategies for The Reproducibility Crisis in AI}

\begin{figure}[t]
    \centering
    \includegraphics[width=0.72\linewidth]{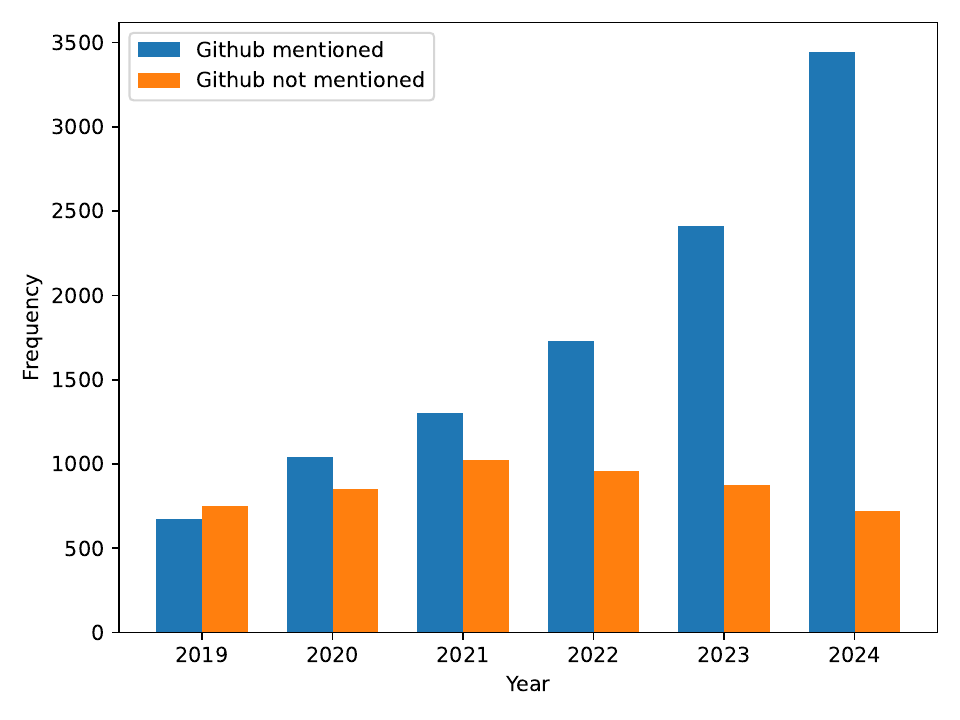}
    \caption{Number of NeurIPS publications that mention GitHub between 2019-2024. We use GitHub mentions as a proxy for replicability. The motivations and limiations of this approach are described in Appendix Section~\ref{sec:GitHub_NeurIPS}}
    \label{fig:NeurIPS_code}
\end{figure}

In recent years a few prominent conferences have engaged in practices to increase reproducibility with examples including the pre-registration workshop at NeurIPS in 2020~\cite{Prereg2020}  and 2021, and the Reproducibility Challenge which has run since 2018~\cite{Pineau}. However, these represent small initiatives with few submissions compared to the main conferences and largely there is no consensus on how to address reproducibility in AI. In Figure~\ref{fig:NeurIPS_code} we present the number of papers at NeurIPS that mention GitHub. We use this as a proxy for replicability of papers, while this is an imperfect measure as discussed in Appendix Section~\ref{sec:GitHub_NeurIPS}, it can be observed that the trend of mentions has increased over the past five years. The most notable changes occurs when NeurIPS introduced the Datasets and Benchmarks track in 2021. The results indicates that more papers are sharing code bases, however, there exists a large number of papers over the last five years that do not provide any mention of code bases. Furthermore, simply replicating a result does not mean that its findings are reproducible in nature, so while this is indicative of reproducibility it is a limited analysis. These findings somewhat support the belief that there is a reproducibility crisis in the field that will slow progress and propagate harm~\cite{ball2023ai}. Given the broad adoption of AI and its growing importance across domains it is of the utmost importance that reproducibility protocols are strengthened.

Of particular relevance when considering practical steps to improve reproducibility standards in AI are \textbf{preregistration, improved statistical experimental design} and finally \textbf{negative result reporting}. Each of these pragmatic reproducibility protocols can greatly improve the information environment and have shown success in other scientific domains. We discuss how leveraging policy can improve reproducibility and enable effective governance outcomes, this is visualised in Figure~\ref{fig:newtheory_of_change} and discussed in Appendix Section~\ref{app:Theory_of_Change}.  

\begin{figure}[H]
    \centering
\includegraphics[width=0.72\linewidth]{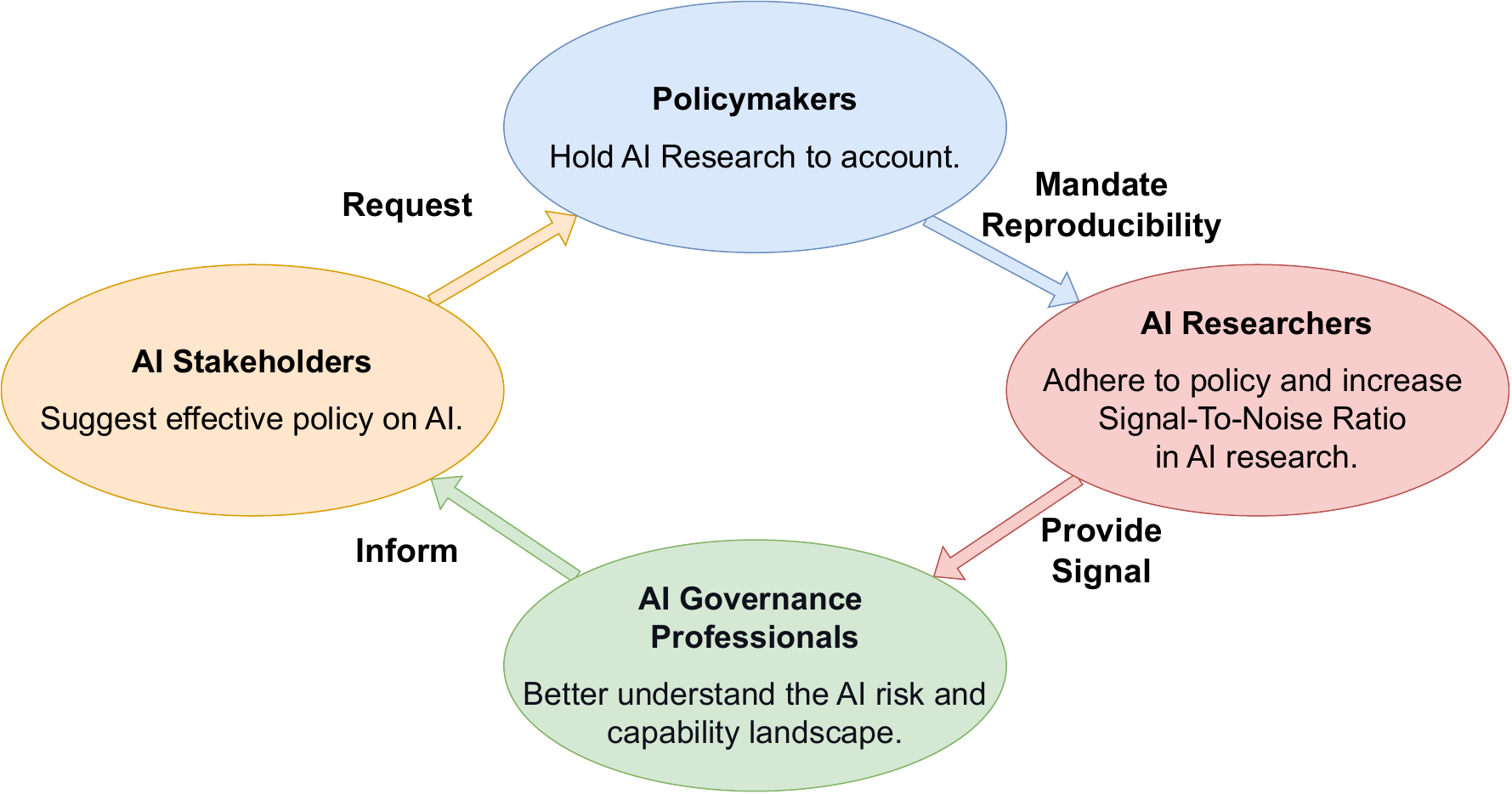}
    \caption{Preposed impact of reproducibility protocols.}
    \label{fig:newtheory_of_change}
\end{figure}

\subsection{Preregistration}
Postdiction can occur in science when a researcher forms a new, or improved, hypothesis to explain their results due to the observation of new data; it represents a typical manifestation of hindsight bias~\cite{roese2012hindsight}. Reliance on such postdictions can lead to overconfidence and inflate the likelihood of false positive results. As a result, this can hinder scientific progress as presenting postdictions as predictions reduces the communication of uncertainties which harms reproducibility~\cite{nosek2018preregistration}. Preregistration has been introduced across domains to improve research standards and strengthen peer review mechanisms. Preregistration enables reviewers to distinguish between prediction and postdiction as the predictions and corresponding experiments to test the hypothesis are publicly registered before the experiment is conducted.~\cite{nosek2018preregistration}. However, it should be noted that while researchers consider preregistration a net-positive, it can induce more stress and cause longer project durations, but, overall, researchers recommend the practice~\cite{sarafoglou2022survey}. For AI, existing preregistration practices can be adopted from other fields such that researchers and policymakers have better estimations over prediction capacities for AI research.  Currently research venues for AI do not require preregistration of experiments, however due to its central role in communicating uncertainties within research it should be adopted for AI research. We see that a preregistration mandate is a viable avenue for increasing the Signal-To-Noise Ratio of AI research, however this would require agreement from publishing venues and reviewers.

\subsection{Statistical Leverage}
In many research domains such as psychology, clinical trails and biology, studies are dependent upon voluntary participation to conduct experiments, this leads to issues surrounding sample size which can impact the robustness of findings. Typically  AI research (outside of human centred studies) does not have a participation bottleneck, making it possible to conduct experiments using high numbers of samples where greater statistical power can be leveraged, such as the robust analysis conducted in physics~\cite{lyons2013discovering}. Despite this, many research venues do not have requirements or guidance surrounding sample sizes used in studies. Without consensus, many papers employ a varying number of sample sizes or omit reporting altogether which increases uncertainty in findings as small sample sizes are unreliable~\cite{cao2024small}. Furthermore, introducing guidance on sample sizes for AI research and appropriate use of statistics such as Standard Error of the Mean~\cite{belia2005researchers} would enable improved analysis reducing statistical errors. The use of large sample sizes has seen benefits in understanding knowledge transfer phenomenon in neural networks~\cite{mason-williams2024knowledge} and has also been suggested to improve capability reporting for evaluations of LLMs~\cite{miller2024adding}, but can and should be applied to AI research more broadly. Possible governance solutions exist in providing open access compute, GPU and CPU, resources from AI Factories or Gigafactories~\footnote{AI Factories:~\url{https://digital-strategy.ec.europa.eu/en/policies/ai-factories}} that are dedicated to enabling improved statistical power of compute based experiments, to champion statistical significance of findings in AI.

\subsection{Negative Result Reporting}
Scientific domains often suffer from publication bias which favors the reporting of only significant or positive results. This is often due to rejection of studies with negative results, opportunity cost of writing up negative results and lack of citation incentive for negative results~\cite{mlinaric2017dealing}; as well as conflicts with funding which favors positive outcomes~\cite{nair2019publication}. As a result, negative result publication is seldom practiced in AI, this means that AI researchers and  governance experts have limited oversight in the current limitations of AI and our understanding of it. Moreover, this can lead to an over-reliance on positive results which can impact policymaking as a full scientific picture cannot be presented leading to ineffective policy implementation~\cite{sharma2019positive}. Given the predicted influence of AI and its broad range of stakeholders it is crucial that scientists, policymakers and the public demand full transparency on the state-of-play of AI capabilities through the publication of negative results. A notable step in the right direction is represented by the Science for Deep Learning NeurIPS Workshop which actively incentivised a debunking challenge to interrogate common wisdom in the field~\footnote{Debunking Challenge at NeurIPS 2024: \url{https://scienceofdlworkshop.github.io/challenge/}}. Research which publishes negative results in AI~\cite{zhang2016understanding,dinh2017sharp,mcgreivy2024weak,mason-williams2025data} has potential to shift perspectives on long-held beliefs which can lead to innovative approaches. In AI negative results reporting can be supported with increased funding towards workshops at conferences that support such open science initiatives to incentivise the write up of negative results.

\section{Conclusion}
Improving reproducibility standards for AI is central to empowering policymakers to execute meaningful and effective governance mechanisms. Given the potential of AI to revolutionise numerous sectors of society it is of the utmost importance that collective action ensures scientific studies in AI are held to the highest standard to avoid the common pitfalls attributed to empirical science. By increasing awareness and calling for consensus on reproducibility protocols it is possible to increase the Signal-To-Noise Ratio of AI. However without cohesive action to address reproducibility there is a high likelihood that the many harms AI can propagate will be actualised. Thus, it is the collective responsibility of scientists, policymakers and governments alike to address reproducibility as a new frontier in AI governance.

\bibliography{reproducibility}
\bibliographystyle{icml2025}
\newpage
\appendix
\onecolumn
\section{Methodology for Figures}

In this section, we detail the methodology employed to create Figures~\ref{fig:speed_vs_reproducibility} and~\ref{fig:NeurIPS_code}; we would like to highlight that these figures are entirely indicative and are not complete reconstructions of the quantities of interest. In the following subsections, we identify the limitations of our analysis and why they should be considered best attempts at capturing quantities of interest, as discussed in the main body of the paper.

\subsection{Speed of Research and Reproducibility Matrix}
\label{sec:Speed_Reproduce}

\begin{figure}[H]
    \centering
    \includegraphics[width=0.5\linewidth]{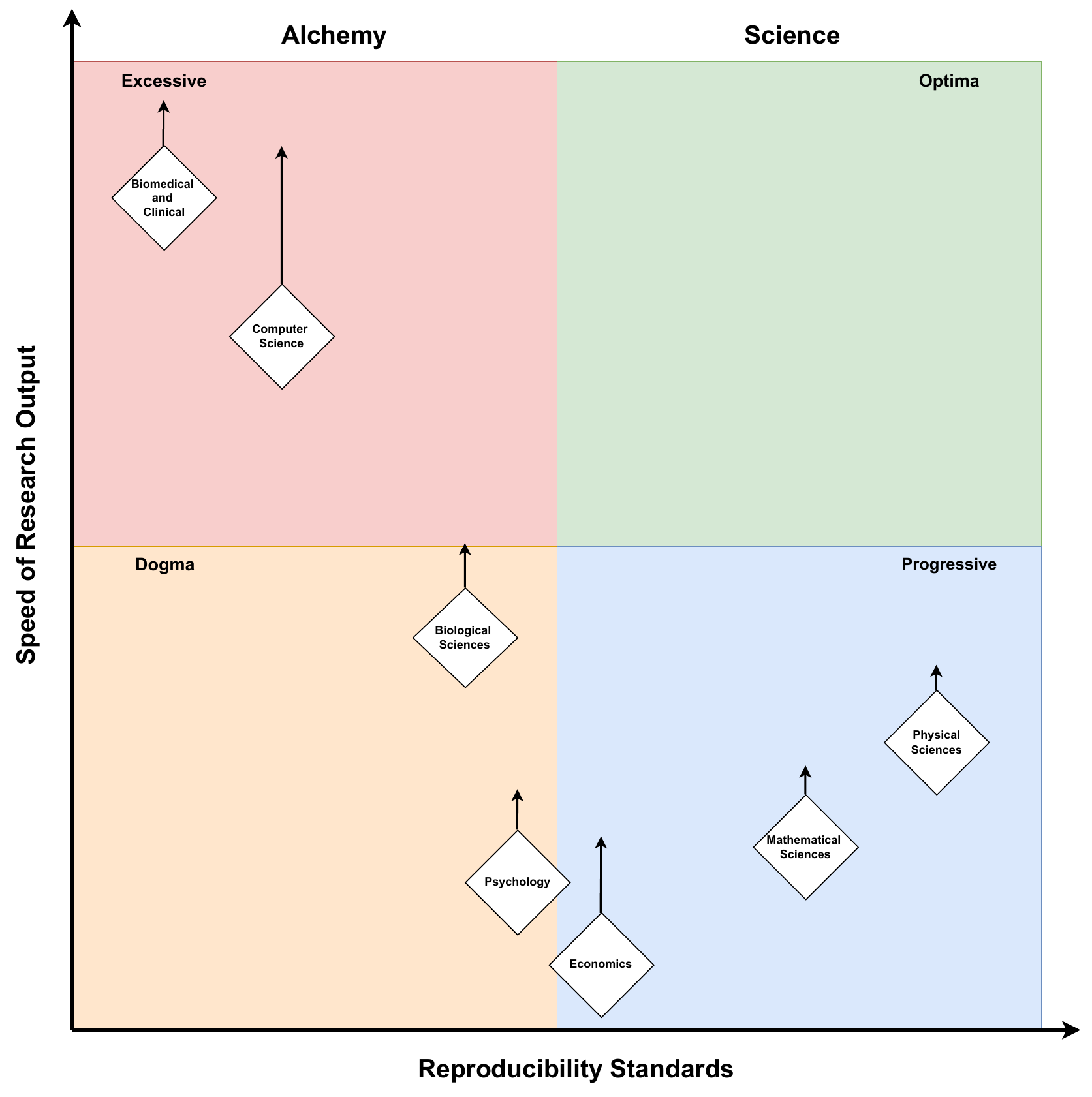}x
    \caption{Blown up plot of publication speed versus reproducibility standards.}
    \label{fig:enter-label}
\end{figure}

\paragraph{Speed of Research Outputs:}
For Figure~\ref{fig:speed_vs_reproducibility}, we use data from the research database Dimensions~\footnote{Research database Dimensions: \url{https://www.dimensions.ai/dimensions-data/}} from the \nth{11} April 2025, which provides publishing analysis of different scientific domains. For the results in the figure, we extract the number of publications recorded for each domain in 2024. Table~\ref{tab:pub_over_years} presents the number of publications for each domain in the last five years between 2024 and 2019. The Information and Computer Science domain representing Artificial Intelligence has the most considerable average year-on-year change, double that of any other domain presented. With this trend set to continue, it underscores the prevalence of research in AI. It shows how important it is to establish strong reproducibility standards for a technology that can be applied across research domains. 

\paragraph{Reproducibility Ratings:} The positions on the reproducibility axis are primarily determined by previous reproduction analysis per domain. For the Biomedical domains previous studies have highlighted that 72\% of surveyed researchers agreed there was a reproducibility crisis in biomedicine, with 27\% stating the crisis was significant~\cite{cobey2024biomedical}, with a recent study in Brazil showing that only 21\% of the experiments were replicable~\cite{R;_2025}. Following this we place Computer Science based on the analysis we conducted and presented in the main body in Table~\ref{tab:pub_over_years} and Figure~\ref{fig:NeurIPS_code}; alongside previous studies which have shown low reproducibility~\cite{gundersen2018state,gundersen2020reproducibility,gundersen2025unreasonable} and repeatability~\cite{collberg2016repeatability} for the Computer Science domain. Additionally, more recent studies have shown low replicability of top-rated papers that gave Oral or Spotlight talks at ICML 2024 with 24\% replication by LLMs at significant computational cost and less than 50\% reproducibility by PhD students~\cite{starace2025paperbench}. A survey on Biological Sciences in 2016 found that 70\% of researchers could not reproduce the findings of other scientists and circa 60\% of researchers could not reproduce their findings~\cite{baker20161}, given this was almost a decade ago and attention has been paid to this we increased its reproducibility score. There are initiatives to improve reproducibility, such as the ASCB Report on Reproducibility and the American Type Culture Collection (ATCC) (Cell and the Microbial Authentication Services and Programs). We believe this is an ongoing issue within Biology, but it is receiving attention from the field. We give Psychology and Economics a moderate reproducibility score due to the implementation of pre-registration practices discussed in the paper's main body. For mathematics, we provide one of the highest reproducibility standards; this is because, by nature of the field, mathematics does not depend on empirical study but rather proof and verifications, which reduces the avenues for error that are observed in more empirical domains~\cite{bordg2021replication}. Finally, we provide the highest reproduction score to the Physical Sciences as it primarily focuses on the creation of theories, and for empirical particle physics experiments, the 5-sigma significance is adopted to ensure exact findings, but with calls to tailor this for the experiment being conducted~\cite{lyons2013discovering}. As a result, physics, namely via particle physics, has the best reproducibility standards due to high significance and strict reporting standards.

\subsubsection{Categories:} We create four categories to describe research outputs based on the speed of outputs and reproducibility ratings. The overarching categories are Alchemy and Science and they are defined in this context as the following. 
\begin{personalstatement}
    \textbf{Alchemy:} Research that largely follows the scientific process but where findings are not reproducible across other settings, or replicable in the original setting, this research is characterised by a \textbf{low Signal-To-Noise Ratio}.
    \\
    \\
    \textbf{Science:} Research that follows the scientific process but where findings are reproducible across other settings and/or replicable in the original setting, this research is characterised by a \textbf{high Signal-To-Noise Ratio}.
\end{personalstatement}

The subcategories are defined as follows:
\begin{personalstatement}
\textbf{Dogma:} This category should be interpreted from the Greek definition of ``something that seems true''; we believe that research domains in this category do not output research quickly but also have low reproducibility standards, which makes them susceptible to dogma.
\\
\\
\textbf{Excessive:} Excessive research is characterised by high-speed research outputs with low reproducibility standards; the research is created quickly, but the findings do not last.
\\
\\
\textbf{Progressive:} Research in this category has a slow publication speed but has high reproducibility standards. In this category progress can be slow-moving but each contribution can have high-impact.
\\
\\
\textbf{Optima:} Characterised by a high publication speed and strong reproducibility practices. Research produced by scientific fields in this category represents the research holy grail where there is no trade-off between the reproducibility of findings and speed of advancement.
\end{personalstatement}

Both subcategories \textbf{Dogma} and \textbf{Excessive} fit into the \textbf{Alchemy} category. We have allocated subcategories \textbf{Progressive} and \textbf{Optima} in the larger \textbf{Science}  category as we argue that the ability to reproduce findings separates alchemy and scientific endeavours.

\subsubsection{Limitations:} The limitations of these results are that we have not conducted an exhaustive analysis across research databases; this may mean that other databases may represent other publication trends; however, we believe that this database largely represents publication trends. Furthermore, we know that AI does not encompass all CS research. However, we recognise it as one of the most active research areas, so we decided that the trends for this research domain would describe trends for AI research. 

\subsection{Replicability Proxy for NeurIPS}
\label{sec:GitHub_NeurIPS}

Below, we provide the code we employed to get the count of papers published at NeurIPS between 2019 and 2024, mentioning GitHub. To have a proxy for replicability, we count the number of accepted papers with GitHub links in the main tracks (2019-2021) and the dataset and benchmark track (2022-2024). It has been argued that having access to code bases can improve the replicability of scientific studies in Computer Science~\cite{gundersen2025unreasonable}. So we feel this is an apt proxy. We limit this analysis to NeurIPS as it is rated as the top publication venue for AI\footnote{Google Scholar rating of NeurIPS: \url{https://scholar.google.co.uk/citations?view_op=top_venues&hl=en&vq=eng_artificialintelligence}}. Furthermore, as replicability is required for reproducibility we believe that this proxy is somewhat indicative for reproducibility. 

\paragraph{Limitations:} It is important to note that our proxy for replicability does not represent the exact number of papers that contain the repositories for their code; it is indeed possible and plausible that papers can contain references to GitHub without providing the code to replicate their work and simply providing code does not guarantee that work can always be replicated let alone their findings reproduced. Finally, our data does not represent all of the papers displayed at NeurIPS, and this count excludes Workshop papers an where the PDF analysis resulted in an error. Furthermore, not all studies are empirical and do not require code links to asses if their work is replicable/reproducible. We view replicability as a prerequisite for reproducibility generally and, as a result, we believe that insights on replicability provide a weak but indicative gauge on the status of reproducibility in AI. 

\textbf{Code to replicate data and figures\footnote{Code for replication: \url{https://github.com/IFMW01/reproducibility-the-new-frontier-in-ai-governance}}}

\section{Theory of Change}

\begin{figure}[H]
\centering
 \subfigure[Current Information Flow.] {\includegraphics[width=0.49\textwidth]{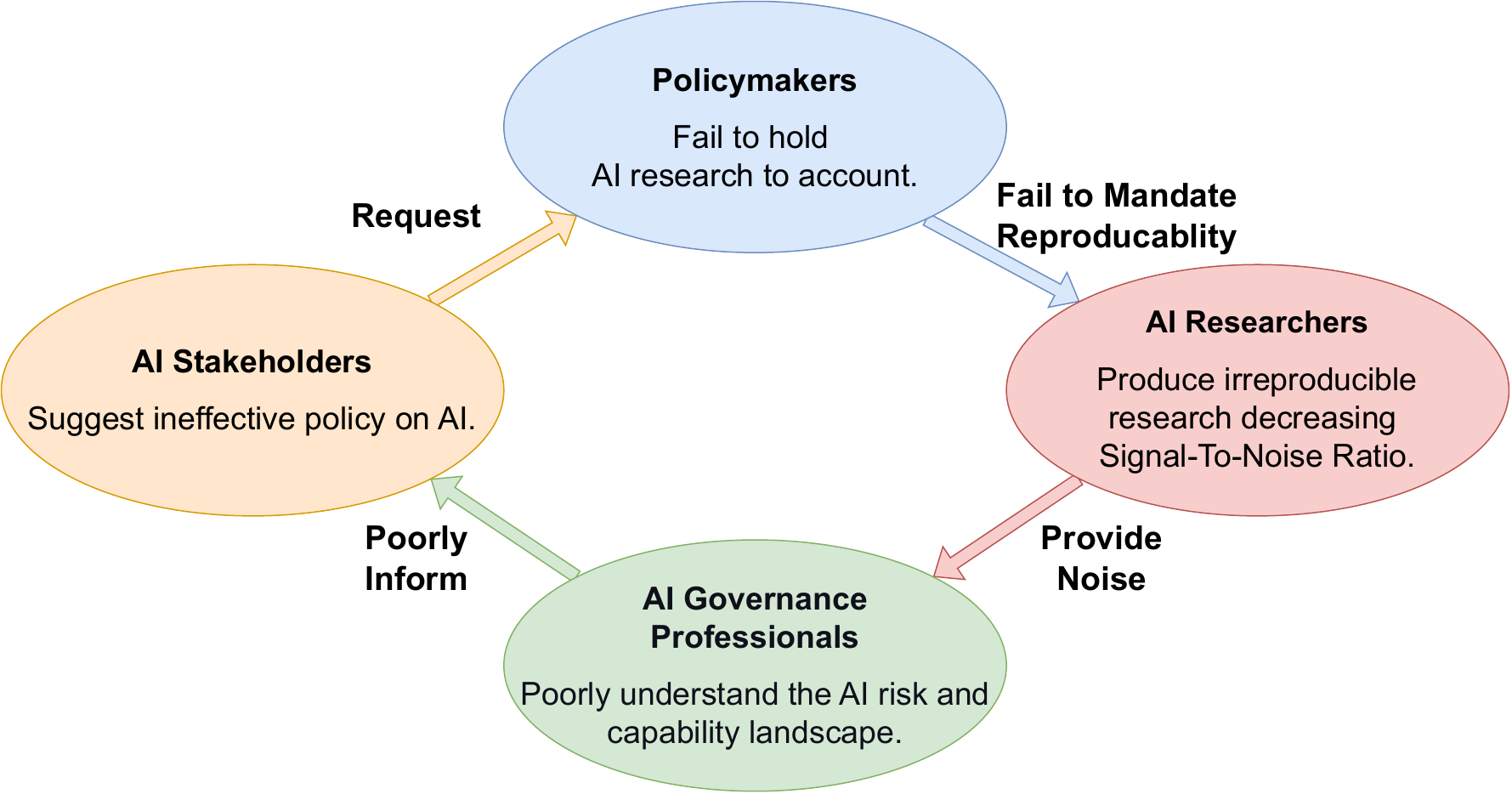}
 \label{current}} 
 \subfigure[Proposed Information Flow.] {\includegraphics[width=0.49\textwidth]{theory_of_change.drawio.pdf}\label{future}} \hfill
 \caption{Reproducibility: Theory of Change.}\label{fig:theory_of_change}
\end{figure}

Currently, irreproducible findings in AI research act as a bottleneck to effective governance. We visualise this in Sub Figure~\ref{current}; here, we argue that Policymakers fail to hold AI research to account because they fail to mandate better standards in AI research, which in turn leads to a decreased Signal-To-Noise-Ratio of outputs, which provides Governance Professionals a more complex challenge in understanding the AI capability and risk landscape, leading to poorly informed AI Stakeholders who suggest policy that will be ultimately ineffective for holding AI research, or AI in general, to account. In Sub Figure~\ref{future}, we show how our suggestions could improve this feedback loop. Introducing mandates of reproducibility standards on AI research, or at least research used for policy, can increase the Signal-To-Noise-Ratio of research outputs, which will provide an improved signal to Governance Professionals who will have a stronger understanding of the capability and risk landscape, which can be effectively communicated to AI stakeholders, such that effective policy on AI is championed. While there are other methods for increasing research standards that research publishers and conferences can take, we believe collaboration between policymakers and these bodies could be most effective at enacting this change. It is important to note that increased reproducibility protocols do come with trade-offs that can slow down research projects, increase burdens for researchers and require more compute (GPU or CPU hours). However, due to the projected impact of AI, we believe this is a necessary step to create a healthy information environment for AI characterised by a high Signal-To-Noise Ratio, which will, in turn, improve governance endeavours. 
\label{app:Theory_of_Change}

\end{document}